\documentclass[runningheads]{llncs}

\usepackage{tikz}

\usepackage[accsupp]{axessibility}  

\usepackage{graphicx}
\usepackage{indentfirst}

\usepackage{comment}
\usepackage{amsmath,amssymb} 
\usepackage{color}
\usepackage{graphicx}
\usepackage{amsmath}
\usepackage{amssymb}
\usepackage{amstext}
\usepackage{booktabs}
\usepackage{multirow}
\usepackage{caption}
\usepackage{subcaption}
\usepackage{threeparttable}
\usepackage[linesnumbered, ruled]{algorithm2e}
\usepackage{wrapfig}
\usepackage{makecell}
\usepackage{epsfig}
\usepackage{url}
\usepackage{flushend}
\usepackage{colortbl}

\definecolor{ggray}{gray}{0.85}
\newcolumntype{a}{>{\columncolor{ggray}}c}
\usepackage[accsupp]{axessibility}  

\usepackage[pagebackref,breaklinks,colorlinks]{hyperref}
\usepackage[capitalize]{cleveref}

\begin{document}
\pagestyle{headings}
\mainmatter
\def\ECCVSubNumber{3409}  

\title{Mimic Embedding via Adaptive Aggregation: Learning Generalizable Person Re-identification} 


\titlerunning{Mimic Embedding via Adaptive Aggregation}
%
\author{Boqiang Xu\inst{1,2} \and
Jian Liang\inst{1,2} \and
Lingxiao He\inst{3} \and Zhenan Sun$^*$ \inst{1,2}}

\authorrunning{Xu et al.}
%
\institute{School of Artificial Intelligence, University of Chinese Academy of Sciences \and
Center for Research on Intelligent Perception and Computing, National Laboratory of Pattern Recognition, Institute of Automation, Chinese Academy of Sciences \and  Longfor Inc. \\ \email{boqiang.xu@cripac.ia.ac.cn}\\\email{\{liangjian92, xiaomingzhidao1\}@gmail.com}\\\email{znsun@nlpr.ia.ac.cn}\\ 
$^*$Corresponding author}

\maketitle

\begin{abstract}
Domain generalizable (DG) person re-identification (ReID) aims to test across unseen domains without access to the target domain data at training time, which is a realistic but challenging problem. 
In contrast to methods assuming an identical model for different domains, Mixture of Experts (MoE) exploits multiple domain-specific networks for leveraging complementary information between domains, obtaining impressive results. 
However, prior MoE-based DG ReID methods suffer from a large model size with the increase of the number of source domains, and most of them overlook the exploitation of domain-invariant characteristics. To handle the two issues above, this paper presents a new approach called Mimic Embedding via adapTive Aggregation ($\mathsf{META}$) for DG person ReID. 
To avoid the large model size, experts in $\mathsf{META}$ do not adopt a branch network for each source domain but share all the parameters except for the batch normalization layers.
Besides multiple experts, $\mathsf{META}$ leverages Instance Normalization (IN) and introduces it into a global branch to pursue invariant features across domains.
Meanwhile, $\mathsf{META}$ considers the relevance of an unseen target sample and source domains via normalization statistics and develops an aggregation module to adaptively integrate multiple experts for mimicking unseen target domain. 
Benefiting from a proposed consistency loss and an episodic training algorithm, $\mathsf{META}$ is expected to mimic embedding for a truly unseen target domain. 
Extensive experiments verify that $\mathsf{META}$ surpasses state-of-the-art DG person ReID methods by a large margin. Our code is available at \href{https://github.com/xbq1994/META}{https://github.com/xbq1994/META}.

\end{abstract}

\section{Introduction}
\label{sec:intro}

Person re-identification (ReID) aims at retrieving persons of the same identity across non-overlapping cameras. 
Many prior works \cite{MGN,miao2019pose,xu2020black,liu2021watching,xu2022learning,he2018deep} have been devoted to the fully-supervised ReID task. 
Despite the promising performance when training and testing on the same domain, the performance always drops significantly when testing on an unseen domain because of the domain shift \cite{msmt}.
To avoid this, recent efforts are devoted to domain adaptive (DA) ReID \cite{UDA1,UDA2,UDA3} and domain generalizable (DG) ReID \cite{metalearning,metanorm,multiexpert,jin2020style}. In contrast to DA ReID, DG ReID is more practical and challenging as it utilizes training data from multiple source domains and directly tests across different and unseen domains, without any target data for training or fine-tuning. In this paper, we mainly focus on the challenging DG person ReID problem. 

\begin{figure}[!t]
\centering
\subcaptionbox{Prior MoE-based DG ReID Method\label{subfig:intro2}}
    {%
        \includegraphics[width=\linewidth]{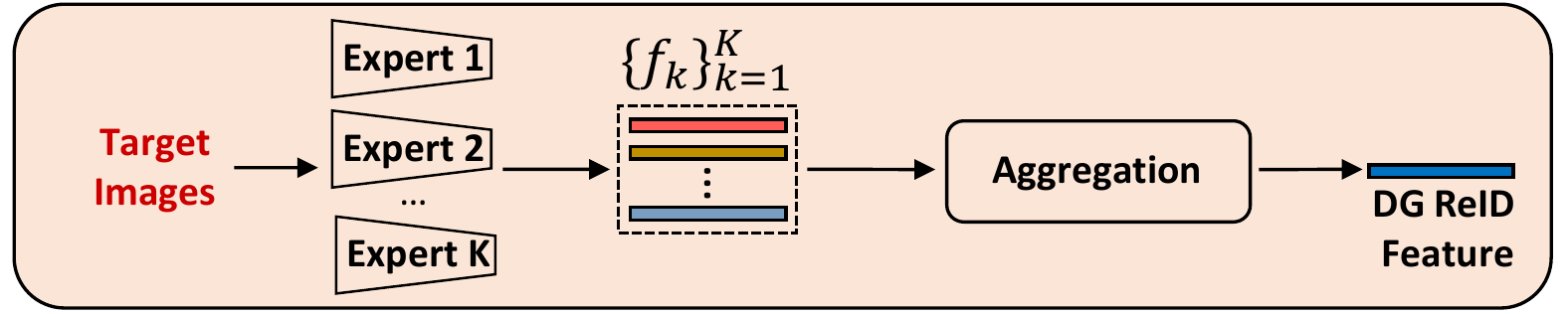}}
\subcaptionbox{Our Method ($\mathsf{META}$)\label{subfig:intro3}}
    {%
        \includegraphics[width=\linewidth]{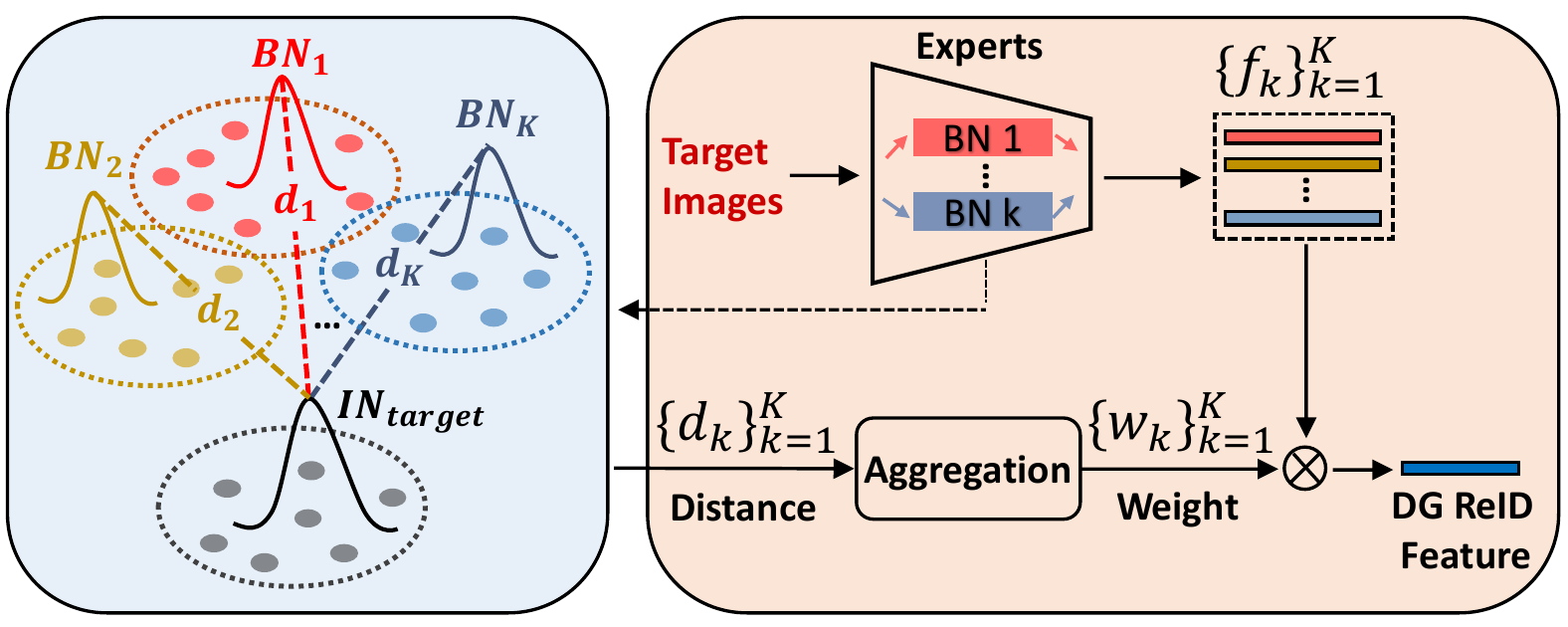}}
\caption{Differences between prior MoE-based DG ReID method and our method. (a) Prior MoE-based DG ReID methods add an individual network (expert) for each source domain, suffering from a large model size with the increase of the number of source domains. (b) Experts in our method share all the parameters except for the batch normalization layers. 
In the testing stage, we calculate the distance between IN statistics of test samples and the BN statistics of source domains for measuring the relevance of target samples w.r.t. source domains. Such distances $\{d_k\}_{k=1}^K$ are exploited by an aggregation module to adaptively integrate multiple experts.}
\end{figure}


Most of the prior DG ReID methods \cite{metalearning,metanorm,DIMN,bai2021person30k,jin2020style} assume an identical model for different domains. However, such an assumption learns a common feature space for different source domains, which may neglect the individual domains' discriminative information and ignore the relevance of the target domain w.r.t source domains. To handle the issues above, mixture of experts (MoE) \cite{jacobs1991adaptive} has been studied for DG ReID, as shown in Fig.~\ref{subfig:intro2}. 
MoE can improve the generalization of models by integrating multiple domain-specific expert networks with the target domain’s inherent relevance w.r.t. diverse source domains.
Generally, prior MoE-based DG ReID methods have two potential problems: 
1) 
As each source domain contains an individual branch network, the model size becomes fairly large with the increase of the number of source domains, limiting the practical deployment.
2) 
Most prior MoE-based DG ReID methods merely focus on learning domain-specific representations but overlook the domain-invariant characteristics.

To tackle the two issues above, 
we propose a novel DG ReID approach called Mimic Embedding via adapTive Aggregation ($\mathsf{META}$), as shown in Fig.~\ref{subfig:intro3}. 
Batch Normalization (BN) statistics are computed on-the-fly during training and can be seen as statistics of the characteristics of individual domain \cite{dg3}. Inspired by this, instead of adding a branch network for each source domain, we train the $\mathsf{META}$ as a lightweight ensemble of multiple experts sharing all the parameters except for the domain-specific BN layers ($i.e.$, one for each source domain for collecting domain-specific BN statistics). By doing so, $\mathsf{META}$ is able to exploit the diversified characteristics of each source domain and meanwhile, keeping the model size from increasing as the source domain increases. 
To extract the domain-invariant features, we design a global branch and leverage Instance Normalization (IN) \cite{instance}, which works as a style normalization layer for filtering out domain-specific contrast information, to explicitly extract domain-invariant features.

Specifically, in our $\mathsf{META}$ method, we exploit individual domains' discriminative information by domain-specific BN layers. 
Then, during testing, the characteristics of the test samples from the unseen domain can be indicated by the means of their IN statistics. 
By measuring the distance between the IN statistics of the test samples and the BN statistics of source domains, we can infer the relevance of the target samples w.r.t. source domains. 
Taking the relevance as input, we further devise a small aggregation module to integrate multiple experts for obtaining the accurate representation of the target person from an unknown domain. By doing so, those relevant source  domains  are  able  to  contribute more valuable information than those less relevant domains.
Moreover, we adopt episodic training \cite{eplearning} which simulates the test process at training time for updating the aggregation module. 
For each training batch, we collect training samples from the same source domain ($e.g.$, $D_k$) to simulate the `unseen target data' for other domain experts. 
We propose a consistency loss to push the aggregated features of other domain experts as discriminative as the features extracted by the expert of $D_k$. In this way, the aggregation module is learned to be able to adaptively integrate diverse domain experts for explicitly mimicking any unseen target domain. 


Our major contributions can be summarized as follows:
\begin{itemize}
 \item We propose $\mathsf{META}$, a novel method to handle the DG ReID problem. Specifically, $\mathsf{META}$ leverages the domain-specific BN layers and designs a global branch to respectively tackle the two issues ($i.e.$, model scalability and oversight in domain invariance) in prior MoE-based DG ReID methods.

 \item We develop a learnable aggregation module, updated by a proposed consistency loss and an episodic training algorithm, to adaptively integrate diverse domain experts via normalization statistics for mimicking any unseen target domain.
 \item Extensive experiments demonstrate that $\mathsf{META}$ surpasses state-of-the-art DG ReID methods by a large margin under various protocols.

\end{itemize}

\section{Related Work}
\textbf{Domain Generalizable Person Re-identification.} Person ReID has made great progress in recent years. Many methods \cite{partbased1,MGN,posebased1,posebased2,he2021semi} have been proposed to improve the ReID performance. Despite the promising performance brought by these methods when training and testing on the same domain, the performance always drops significantly when testing on an unseen domain because of the domain shift \cite{msmt}. To tackle this problem, some researchers start to study the unsupervised domain adaption (UDA) methods \cite{UDA1,UDA2,UDA3}. However, UDA requires unlabeled data from the target source, which is sometimes difficult to be collected in practical applications. As a result, domain generalizable (DG) ReID \cite{metalearning,metanorm,multiexpert,jin2020style} have captivated researchers recently. Generally, DG ReID utilizes training data from multiple source domains and directly tests across different and unseen domains, without any target data for training or fine-tuning. 

We briefly classify prior DG ReID methods into three categories. The first category is Meta-Learning \cite{metalearning,metanorm,DIMN,bai2021person30k}. Meta-learning is a training strategy, which adopts the concept of 'learning to learn' by exposing the model to domain shift during training for learning more generalizable models. Zhao $et$ $al.$ \cite{metalearning} proposed a Memory-based Multi-Source Meta-Learning (M$^3$L) framework, which overcomes the unstable meta-optimization by a memory-based and non-parametric identification loss.

The second category is Domain Alignment \cite{jin2020style}, which attempts to minimize the differences between source domains for pursuing the invariant features across domains. Jin $et$ $al.$ \cite{jin2020style} propose a Style Normalization and Restitution (SNR) module to separate the identity-relevant and identity-irrelevant features by a dual causality loss constraint.

The third category is Mixture of Experts (MoE) \cite{multiexpert}.  MoE learns diverse experts for different domains and takes the target domain’s inherent relevance w.r.t. diverse source domains into consideration for better generalization. Dai $et$ $al.$ \cite{multiexpert} proposed a method called the relevance-aware mixture of experts (RaMoE), which adds a branch network (expert) for each source domain, and designs a voting network for integrating multiple experts. However, \cite{multiexpert} suffers from 
a large model size with the increase of the number of source domains, which limits the application of the RaMoE. To tackle this problem, experts in our method share all the parameters except for the batch normalization layers.

\textbf{Domain-Specific Batch Normalization.} The statistics of BN vary in different domains. Therefore, mixing multiple source domains' statistics may be detrimental to improving generalizable performance \cite{zhou2021domain}. To tackle this problem, domain-specific BN has been studied recently \cite{dg1,dg2,dg3,dg4}. Domain-specific BN works as constructing domain-specific classifiers but shares most of the parameters except for the BN layers. 


\begin{figure*}[t]
 \centering
 \includegraphics[width=\textwidth]{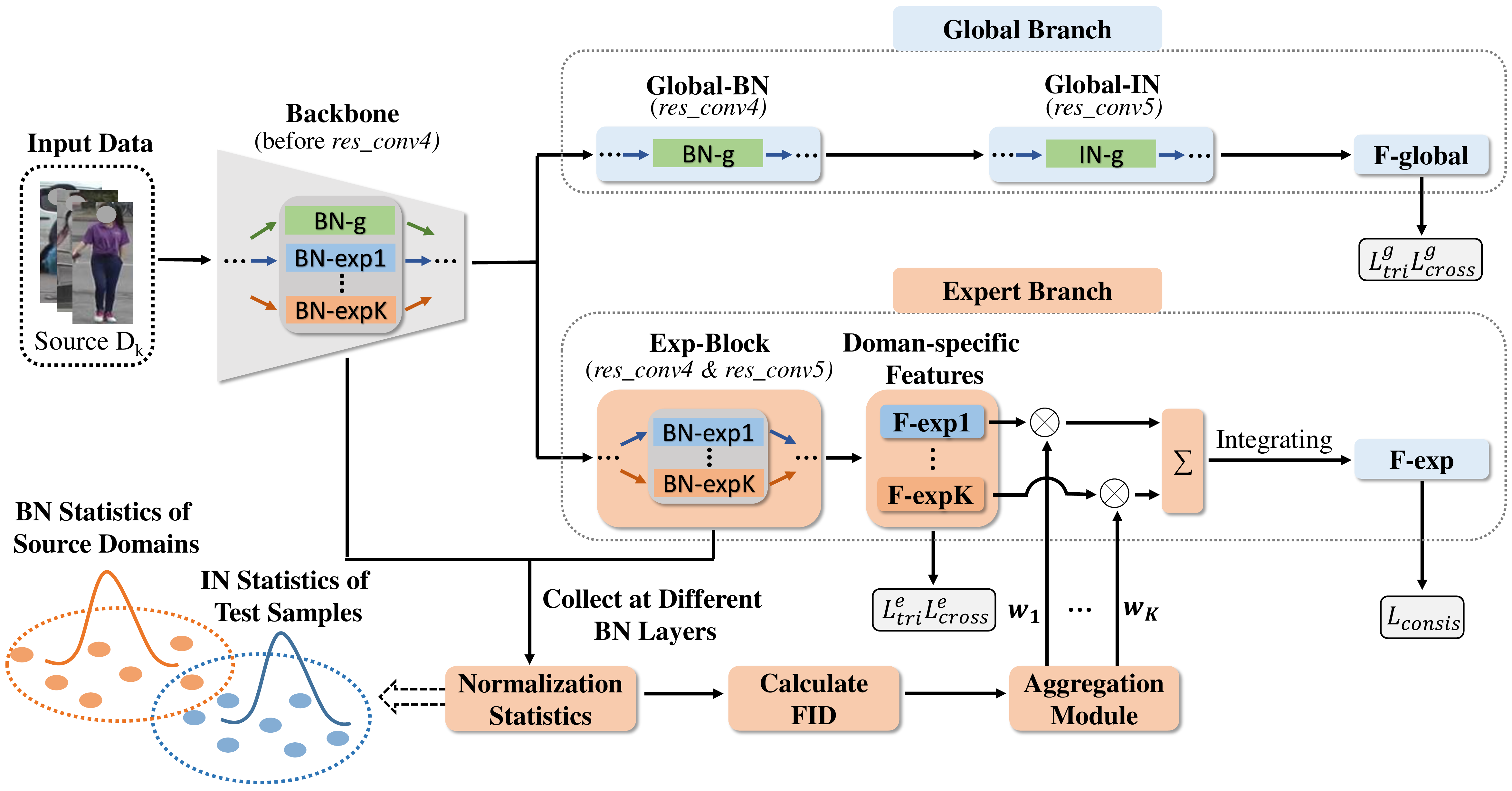}
 \caption{Overview of the proposed $\mathsf{META}$. `$\otimes$' is the operation of element-wise multiplication. `$\sum$' is a series of features' operation: element-wise division or summation. $\mathsf{META}$ is composed of a global branch for capturing domain-invariant features and an expert branch for exploiting complementary domain-specific information. The Exp-Block contains $K$ domain-specific BN layers while the backbone contains $K$ domain-specific BN layers and a global layer BN-g. We replace the BN layers in the $res\_conv5$ with IN layers to construct Global-IN. K domain-specific BN layers are updated by their corresponding source domain's data to capture domain-specific characteristics while BN-g and IN-g are updated by the training data from all the source domains to help extract domain-invariant features. In the expert branch, we collect the IN statistics of the test samples and BN statistics of the source domains at different BN layers and calculate the \emph{Fréchet Inception Distance} (FID) between them to measure the relevance of target samples w.r.t. source domains. Such relevance is leveraged by an aggregation module to adaptively integrate multiple experts. Finally, we concatenate F-global and F-exp for inference.}
 \label{fig:model}
\end{figure*}

\section{Methodology}
Typically, we are provided with $K$ source domains $\left\{D_k \right\}_{k=1}^K$ for training a DG ReID model, which have completely disjoint label spaces. In the testing phase, we directly test on unseen target domains without additional model updating. The structure of the $\mathsf{META}$ is illustrated in Fig.~\ref{fig:model}.

\subsection{Preliminary}
In almost all the prior DG ReID methods \cite{metalearning,metanorm,DIMN,bai2021person30k}, they share BN layers for all the source domains, which may neglect individual domains’ discriminative characteristics and be detrimental to dealing with the domain gap \cite{multiexpert,2019Domain}. To leverage the complementary information of the source domains, inspired by \cite{dg3,2019Domain,bai2021unsupervised}, we adopt \emph{domain-specific batch normalization} in $\mathsf{META}$. 

Let $X_k \in \mathbb{R}^{N \times C \times H \times W} $ denotes a feature map extracted from source domain $D_k$, where $N,C,H,W$ respectively indicate the batch size, the number of channels, the height, and the width. BN layer normalizes features by:
\begin{equation}
BN(X_k)=\gamma_k^{bn}\cdot\frac{X_k-\mu_k^{bn}}{\sqrt{{\sigma_k^{bn}}^2+\epsilon}}+\beta_k^{bn}, 
\label{equ:bn}
\end{equation}
where $\gamma_k^{bn} \in \mathbb{R}^C$ and $\beta_k^{bn} \in \mathbb{R}^C$ are affine parameters, $\epsilon>0$ is a small constant to avoid divided-by-zero.
$\mu_k^{bn} \in \mathbb{R}^C$ and $\sigma_k^{bn} \in \mathbb{R}^C$ are respectively mean value and standard deviation calculated with respect to a mini-batch and each channel:
\begin{equation}
\mu_k^{bn}=\frac{\sum\nolimits_{n}\sum\nolimits_{h,w}X_k}{N\cdot H \cdot W} \quad\text{and}\quad \sigma_k^{bn}=\sqrt{\frac{\sum\nolimits_{n}\sum\nolimits_{h,w}(X_k-\mu_k^{bn})^2}{N\cdot H \cdot W}}.
\label{equ:bn1}
\end{equation}
$\mu_k^{bn}$ and $\sigma_k^{bn}$ are updated by the moving average operation \cite{ioffe2015batch} at training time and fixed during inference. We design individual BN layers for each source domain. Specifically, as shown in Fig.~\ref{fig:model}, Exp-Block contains $K$ domain-specific BN layers, which are updated by the training data from the corresponding source domain to exploit domain-specific characteristics. 
Besides $K$ domain-specific BN layers, another global layer BN-g is introduced in the backbone and global branch, which is updated by the training data from all the source domains to help extract domain-invariant features. 

Although we have exploited the complementary information of the source domains via \emph{domain-specific batch normalization}, it is still challenging to approximate the population statistics of the unseen target domain because target domain data cannot be accessed at training time. To do this, at testing time, we rely on IN statistics to capture the characteristics of the target samples. Given an example from target domain $T_t$, IN layers normalize features by:
\begin{equation}
IN(X_t)=\gamma_t^{in}\cdot\frac{X_t-\mu_t^{in}}{\sqrt{{\sigma_t^{in}}^2+\epsilon}}+\beta_t^{in}. 
\label{equ:in}
\end{equation}
Different from BN, mean value $\mu_t^{in}$ and standard deviation $\sigma_t^{in}$ here are calculated with respect to each sample and each channel:
\begin{equation}
\mu_t^{in}=\frac{\sum\nolimits_{h,w}X_t}{H \cdot W} \quad\text{and}\quad \sigma_t^{in}=\sqrt{\frac{\sum\nolimits_{h,w}(X_t-\mu_t^{in})^2}{H \cdot W}}.
\label{equ:in1}
\end{equation}

In the next section, we explain how to measure the relevance of the target samples w.r.t. source domains via BN and IN statistics.

\subsection{Expert Branch in $\mathsf{META}$}
We expect those relevant source  domains  to  contribute more valuable information than those less relevant domains.
In this section, we explain how to measure the relevance of the target samples w.r.t. source domains via BN and IN statistics for integrating multiple experts. 
From Eq.~(\ref{equ:bn})-Eq.~(\ref{equ:in1}), we can see that IN is the degenerate case of BN with batch size $N$ equal to 1. 
$\mathsf{META}$ is built on such observation that BN and IN statistics are both approximations of Gaussian distributions ($i.e.,$ they are comparable) and have potential to reflect the properties of the source domains and target samples respectively. Therefore, we can measure the relevance of the target samples w.r.t. source domains by comparing IN and BN statistics of them.

Specifically, we collect the BN statistics of source domains at different BN layers. Considering a source domain $D_k$, we denote $D_k^{(l)}=(\mu_k^{bn(l)},\sigma_k^{bn(l)^2})$ the BN statistics at $l$-th layer of $k$-th BN-exp. For each test sample $x_t$ from an unseen target domain $T_t$, we forward propagate $x_t$ through the network and calculate its IN statistics by Eq.~(\ref{equ:in1}) at $l$-th layer of $k$-th BN-exp as $T_t^{(l)} = (\mu_t^{in(l)},\sigma_t^{in(l)^2})$. 
We adopt \emph{Fréchet Inception Distance} (FID) \cite{heusel2017gans} to compute the distance between the BN and IN statistics at $l$-th layer as: 
\begin{equation}
\begin{aligned}
r_{k,t}^{(l)}&=FID((\mu_k^{bn(l)},\sigma_k^{bn(l)^2}), (\mu_t^{in(l)},\sigma_t^{in(l)^2}))\\
&= \|\mu_k^{bn(l)}-\mu_t^{in(l)}\|_2^2 + Tr(C_k^{(l)}+C_t^{(l)}-2(C_k^{(l)}C_t^{(l)})^{\frac{1}{2}}),\\
\text{where}\; C_k^{(l)}&= Diag(\sigma_k^{bn(l)^2}),\; C_t^{(l)}= Diag(\sigma_t^{in(l)^2}),  
\end{aligned}
\end{equation}
and $Diag(\cdot)$ returns a square diagonal matrix with the elements of input vector on the main diagonal. $r_{k,t}^{(l)}$ denotes the distance between the BN statistics of source domain $D_k$ and IN statistics of test sample from target domain $T_t$ at $l$-th layer, $\mid\mid\cdot\mid\mid$ denotes the Euclidean norm, and $Tr(\cdot)$ denotes the trace of the matrix. Thereafter, we concatenate $r_{k,t}^{(l)}$ at every layer as:
\begin{equation}
R_k^t=[r_{k,t}^{(1)},r_{k,t}^{(2)},...,r_{k,t}^{(L)}] \in \mathbb{R}^{1 \times L}.
\end{equation}

Then, we forward propagate $R_k^t$ to an aggregation module $h: \mathbb{R}^{L}\to \mathbb{R}$ for computing the weight of domain-specific expert: 
\begin{equation}
w_k = h(R_k^t),
\end{equation}
where $h$ consists of two fully-connected layers. The aggregation module further enhances the domains' relevance measure by adopting a learnable module. 
During testing, we get the \emph{F-exp} as a linear combination of the multiple experts:
\begin{equation}
\emph{F-exp}(x) = \sum\limits_{k=1}^{K}\frac{e^{w_k}f(x \mid k)}{\sum\nolimits_{j}e^{w_j}}, 
\label{equ:mix}
\end{equation}
where $f(x \mid k)$ is the result of a forward pass of the $k$-th expert in the network. During training, we get the \emph{F-exp} in another way, which will be introduced in Section 3.4. In this way, relevant source domains are able to contribute more valuable information than those less relevant domains for better generalization performance on the target domain.

\subsection{Global Branch in $\mathsf{META}$}
We design a global branch to learn the domain-invariant features, which works as a complement to the domain-specific representations extracted by the expert branch for better generalizability. IN works on normalizing features with the statistics of individual instances, by which the domain-specific information could be filtered out from the content \cite{instance}. Inspired by this, we leverage IN layers in the global branch to capture the domain-invariant features.

The global branch is designed based on the findings from  \cite{pan2018two} that adding IN layers after BN layers could significantly improve the domain generalization performance of the model. Specifically, as shown in Fig.~\ref{fig:model}, the global branch is composed of the \emph{Global-Bn} and \emph{Global-In} blocks. \emph{Global-Bn} block is the same as $res\_conv4$. We replace all the BN layers in the $res\_conv5$ with IN layers to build the \emph{Global-In} block. Furthermore, training samples from all the source domains are used to update the global branch.

\subsection{Training Policy}
At training time, each training batch is composed of the training samples collected from the same source domain. Let $x$ denotes the current training sample collected from source domain $D_i$ ($1 \leq i \leq K$). As shown in Fig.~\ref{fig:model}, we freeze all the BN layers except for the BN-g and $i$-th BN-exp. We update the global branch by the triplet loss \cite{triplet} $\mathcal{L}^g_{tri}$ and cross-entropy loss $\mathcal{L}^g_{cross}$. Meanwhile, we optimize the $i$-th expert by the triplet loss \cite{triplet} $\mathcal{L}^e_{tri}$ and cross-entropy loss $\mathcal{L}^e_{cross}$.
Combining these losses above together, we have the following overall objective:
\begin{equation}
\mathcal{L}_{base} = \mathcal{L}^g_{tri} + \mathcal{L}^g_{cross} + \mathcal{L}^e_{tri} + \mathcal{L}^e_{cross} .
\label{equ:base}
\end{equation}

In addition, we adopt episodic training \cite{eplearning} which simulates the test process at training time to update the aggregation module. When $x$ is input to the network, domain $D_i$ is seemed as the `unseen target domain' to the other $K-1$ domain-specific experts $\left\{f(x \mid k) \right\}_{k=1,k \neq i}^K$. We combine these $K-1$ domain experts to produce the representation \emph{F-exp}, which is formulated as:
\begin{equation}
\emph{F-exp}(x) = \sum\limits_{k=1,k \neq i}^{K}\frac{e^{w_k}f(x \mid k)}{\sum\nolimits_{j,j \neq i}e^{w_j}}, \;x \in D_i,
\label{equ:fexp}
\end{equation}
where $w_k$ is the weight of $k$-th expert and $f(x \mid k)$ is the result of a forward pass of the $k$-th expert. To mimic embedding of $D_i$ with \emph{F-exp}, we propose a consistency loss to push the aggregated feature \emph{F-exp} as discriminative as the feature $f(x \mid i)$ extracted by the $i$-th expert. The consistency loss is formulated as:
\begin{equation}
\mathcal{L}_{consis} = [\alpha_1 + \Gamma^+_{exp}-\Gamma^+_{i}]_+ + [\alpha_2 + \Gamma^-_{i}-\Gamma^-_{exp}]_+ ,
\label{equ:consis}
\end{equation}
where $\alpha_1$ and $\alpha_2$ are margins, $\Gamma^+_{exp}$ and $\Gamma^+_{i}$ are hardest positive distances \cite{triplet} of \emph{F-exp} and $f(x \mid i)$ respectively, $\Gamma^-_{exp}$ and $\Gamma^-_{i}$ are hardest negative distances \cite{triplet} of \emph{F-exp} and $f(x \mid i)$ respectively, $[z]_+$ equals to $max(z,0)$. By minimizing Eq.~(\ref{equ:consis}), the aggregation module is learned to explicitly mimic the target domain via multiple experts. The total loss can be formulated as:
\begin{equation}
\mathcal{L} = \mathcal{L}_{base} + \mathcal{L}_{consis} .
\end{equation} 

\begin{algorithm}[t]
\small
  \SetKwInOut{Input}{Input}\SetKwInOut{Output}{Output}
  
  \KwIn{Training data $x$ from source domain $D_i$; MaxIters; MaxEpochs.}
  \KwOut{Feature extractor $F_\theta (\cdot)$; Domain-specific experts $\left\{f(x \mid k) \right\}_{k=1}^K$.}
  Initialization\;
  \For{epoch=1 \KwTo MaxEpochs}{
    \For{iter=1 \KwTo MaxIters}{
    \textbf{Domain-specific BN layers:}
    
    Freeze all the BN layers except for the BN-g and $i$-th BN-exp\;

    \textbf{Global Branch:}
    
    Update global branch by $\mathcal{L}_{tri}^g$ and $\mathcal{L}_{cross}^g$\;
    
    \textbf{Expert Branch:}
    
    Update expert branch by $\mathcal{L}_{tri}^e$ and $\mathcal{L}_{cross}^e$\;    
    
    \textbf{Aggregation Module:}
    
    Combine $\left\{f(x \mid k) \right\}_{k=1,k \neq i}^K$ by Eq.~(\ref{equ:fexp}) to produce \emph{F-exp}\;
    
    Update aggregation module by $\mathcal{L}_{consis}$ in Eq.~(\ref{equ:consis})\;
    }
  }
  \caption{Training Procedure of $\mathsf{META}$}\label{alg:train}
\end{algorithm}

At test time, we combine $K$ domain experts by Eq.~(\ref{equ:mix}) to produce \emph{F-exp}, and concatenate it with \emph{F-global} as the final representation. The overall training procedure is shown in Algorithm~\ref{alg:train}. 

\section{Experiments}
\subsection{Datasets and Settings}

\textbf{Datsets.} We conduct extensive experiments on 9 public ReID or person search datasets including Market1501 \cite{market}, MSMT17 \cite{msmt}, CUHK02 \cite{cuhk02}, CUHK03 \cite{cuhk03}, CUHK-SYSU \cite{cuhksysu}, PRID \cite{prid}, GRID \cite{grid}, VIPeR \cite{viper}, and iLIDs \cite{ilids}. The details of these datasets are illustrated in Table~\ref{table:datasets}. For CUHK03, we use the 'labeled' data as \cite{multiexpert}. For simplicity, we denote MSMT17 as MS, Market1501 as M, CUHK02 as C2, CUHK03 as C3, and CUHK-SYSU as CS. We utilize Cumulative Matching Characteristics (CMC) and mean average precision (mAP) for evaluation.

\setlength{\tabcolsep}{3.0pt}
\begin{table}[h]
\centering
\begin{minipage}{0.50\linewidth}
\centering
\footnotesize
\caption{Summary of all the datasets.}
\resizebox{0.99\textwidth}{!}{$
\begin{tabular}{lrrc}
\toprule
Datasets & \#IDs & \#Images & \#Cameras \\ 
\midrule
Market1501 (M) \cite{market}& 1,501 & 32,217 & 6 \\
MSMT17 (MS) \cite{msmt} & 4,101 & 126,441 & 15 \\
CUHK02 (C2) \cite{cuhk02} & 1,816 & 7,264 & 10 \\
CUHK03 (C3) \cite{cuhk03} & 1,467 & 14,096 & 2 \\
CUHK-SYSU (CS) \cite{cuhksysu} & 11,934 & 34,574 & 1 \\
PRID \cite{prid} & 749 & 949 & 2 \\
GRID \cite{grid} & 1,025 & 1,275 & 8 \\
VIPeR \cite{viper} & 632 & 1,264 & 2 \\
iLIDs \cite{ilids} & 300 & 4,515 & 2 \\ 
\bottomrule
\end{tabular}
	$}
\label{table:datasets}
\end{minipage}
\begin{minipage}{0.47\linewidth}
\centering
\footnotesize
\caption{Evaluation protocols.}
\resizebox{0.94\textwidth}{!}{$
\begin{tabular}{ccc}
\toprule
& Training Sets & Testing Sets \\ 
\midrule
Protocol-1 & Full-(M+C2+C3+CS) & \begin{tabular}[c]{@{}c@{}}PRID,GRID,\\ VIPeR,iLIDs\end{tabular} \\ 
\midrule
\multirow{3}{*}{Protocol-2} & M+MS+CS & C3 \\
 & M+CS+C3 & MS \\
 & MS+CS+C3 & M \\ 
 \midrule
\multirow{3}{*}{Protocol-3} & Full-(M+MS+CS) & C3 \\
 & Full-(M+CS+C3) & MS \\
 & Full-(MS+CS+C3) & M \\ \bottomrule
\end{tabular}
$}
\label{table:protocol}
\end{minipage}
\end{table}

\textbf{Evaluation Protocols.} Because DukeMTMC-reID \cite{duke}, which was widely used in previous work \cite{metalearning,metanorm,DIMN,bai2021person30k} on DG ReID, has been taken down, we set three new protocols for DG ReID, as shown in Table~\ref{table:protocol}. For protocol-1, we use all the images in the source domains ($i.e.$, including training and testing sets) for training. For PRID, GRID, VIPeR, and iLIDS, following \cite{multiexpert}, the results are evaluated on the average of 10 repeated random splits of query and gallery sets. For protocol-2, we choose one domain from M+MS+CS+C3 for testing and the remaining three domains for training. As the CS person search dataset only contains 1 camera, CS is not used for testing. The difference between protocol-2 and protocol-3 is that we use all the images in the source domains for training under protocol-3.

\setlength{\tabcolsep}{1.0pt}
\begin{table}[t]
\center
\setlength{\abovecaptionskip}{0.2cm}
\setlength{\belowcaptionskip}{0.2cm}
\caption{Comparison with state-of-the-art methods under protocol-1. All the images in the source domains are used for training. The illustration of abbreviations is shown in  Table~\ref{table:datasets}. We report some results of other methods which leverage DukeMTMC-reID in the source domains, while we remove DukeMTMC-reID from our training sets. Although we use fewer source domains, we still get the best performance. `$^{\ast}$' indicates that we re-implement this work based on the authors' code on Github. The best (in \textbf{\color{red}{bold red}}), the second best (in \textit{\color{blue}{italic blue}}).}
\resizebox{1\textwidth}{!}{
\begin{tabular}{lcccccccccaa}
\toprule
\multirow{2}{*}{Method} &  \multirow{2}{*}{\begin{tabular}[c]{@{}c@{}}Source \\ Domains\end{tabular}} & \multicolumn{2}{c}{$\to$PRID} & \multicolumn{2}{c}{$\to$GRID} & \multicolumn{2}{c}{$\to$VIPeR} & \multicolumn{2}{c}{$\to$iLIDs} & \multicolumn{2}{a}{Average} \\  
 & & mAP & Rank-1 & mAP & Rank-1 & mAP & Rank-1 & mAP & Rank-1 & mAP & Rank-1 \\ \midrule
CrossGrad \cite{crossgrad} & \multirow{8}{*}{\begin{tabular}[c]{@{}c@{}}M+D\\ +C2+C3\\+CS\end{tabular}} & 28.2 & 18.8 & 16.0 & 8.96 & 30.4 & 20.9 & 61.3 & 49.7 & 34.0 & 24.6 \\
Agg\_PCB \cite{aggpcb} & & 45.3 & 31.9 & 38.0 & 26.9 & 54.5 & 45.1 & 72.7 & 64.5 & 52.6 & 42.1 \\
MLDG \cite{MLDG}  & & 35.4 & 24.0 & 23.6 & 15.8 & 33.5 & 23.5 & 65.2 & 53.8 & 39.4 & 29.3 \\
PPA \cite{PPA} & & 32.0 & 21.5 & 44.7 & 36.0 & 45.4 & 38.1 & 73.9 & 66.7 & 49.0 & 40.6 \\
DIMN \cite{DIMN} & & 52.0 & 39.2 & 41.1 & 29.3 & 60.1 & 51.2 & 78.4 & 70.2 & 57.9 & 47.5 \\
SNR \cite{jin2020style} & & 66.5 & 52.1 & 47.7 & 40.2 & 61.3 & 52.9 & \textit{\color{blue}89.9} & \textit{\color{blue}84.1} & 66.4 & 57.3 \\
RaMoE \cite{multiexpert} & & 67.3 & 57.7 & 54.2 & 46.8 & 64.6 & 56.6 & \textbf{\color{red}{90.2}} & \textbf{\color{red}{85.0}} & 62.0 & \textit{\color{blue}61.5} \\
DMG-Net \cite{bai2021person30k} & & 68.4 & 60.6 & 56.6 & \textit{\color{blue}51.0} & 60.4 & 53.9 & 83.9 & 79.3 & 67.3 & 61.2 \\ \midrule
QAConv$_{50}$ \cite{qaconv}$^{\ast}$ & \multirow{4}{*}{\begin{tabular}[c]{@{}c@{}}M\\ +C2+C3\\+CS\end{tabular}} & 62.2 & 52.3 & 57.4 & 48.6 & 66.3 & 57.0 & 81.9 & 75.0 & 67.0 & 58.2 \\
M$^{3}$L(ResNet-50) \cite{metalearning}$^{\ast}$ & & 65.3 & 55.0 & 50.5 & 40.0 & \textit{\color{blue}68.2} & \textit{\color{blue}60.8} & 74.3 & 65.0 & 64.6 & 55.2 \\
MetaBIN \cite{metanorm}$^{\ast}$ & &\textit{\color{blue}70.8} &\textit{\color{blue}61.2} &\textit{\color{blue}57.9} &50.2 &64.3 &55.9 &82.7 &74.7 &\textit{\color{blue}68.9} &60.5  \\
$\mathsf{META}$ & & \textbf{\color{red}{71.7}} & \textbf{\color{red}{61.9}} & \textbf{\color{red}{60.1}} & \textbf{\color{red}{52.4}} & \textbf{\color{red}{68.4}} & \textbf{\color{red}{61.5}} & 83.5 & 79.2 & \textbf{\color{red}{70.9}} & \textbf{\color{red}{63.8}} \\ \bottomrule
\end{tabular}
}
\label{table:protocol1}
\end{table}

\setlength{\tabcolsep}{1.0pt}
\begin{table*}[t]
\center
\caption{Comparison with state-of-the-art methods under protocol-2 and protocol-3. `Training Sets' denotes that only the training sets in the source domains are used for training and `Full Images' denotes that all images are leveraged at training time. The illustration of abbreviations is shown in  Table~\ref{table:datasets}. `$^{\ast}$' indicates that we re-implement this work based on the authors' code on Github. The best (in \textbf{\color{red}{bold red}}), the second best (in \textit{\color{blue}{italic blue}}). }
\resizebox{1\textwidth}{!}{
\begin{tabular}{lcccccccaa}
\toprule
\multirow{2}{*}{Method} & \multirow{2}{*}{Setting} &
\multicolumn{2}{c}{\makecell*[c]{M+MS+CS\\$\to$C3}} & \multicolumn{2}{c}{\makecell*[c]{M+CS+C3\\$\to$MS}} & \multicolumn{2}{c}{\makecell*[c]{MS+CS+C3\\$\to$M}} & \multicolumn{2}{a}{Average} \\ 
 & & mAP & Rank-1 & mAP & Rank-1 & mAP & Rank-1 & mAP & Rank-1 \\ \midrule
 SNR$^{\ast}$ \cite{jin2020style} & \multirow{6}{*}{\begin{tabular}[c]{@{}c@{}}Protocol-2\\ (Training Sets)\end{tabular}} & 8.9 & 8.9 & 6.8 & 19.9 & 34.6 & 62.7 & 16.8 & 30.5 \\
 QAConv$_{50}$ \cite{qaconv}$^{\ast}$ & & 25.4 & 24.8 & 16.4 & \textit{\color{blue}{45.3}} & \textit{\color{blue}{63.1}} & \textit{\color{blue}{83.7}} & 35.0 & 51.3 \\
 M$^{3}$L (ResNet-50) \cite{metalearning}$^{\ast}$ & &  20.9 & 31.9 & 15.9 & 36.9 & 58.4 & 79.9 & 31.7 & 49.6 \\
 M$^{3}$L (IBN-Net50) \cite{metalearning}$^{\ast}$ & &  \textit{\color{blue}{34.2}} & \textit{\color{blue}{34.4}} & 16.7 & 37.5 & 61.5 & 82.3 & \textit{\color{blue}{37.5}} & \textit{\color{blue}{51.4}} \\
 MetaBIN \cite{metanorm}$^{\ast}$ & &  28.8 & 28.1 & \textit{\color{blue}{17.8}} & 40.2 & 57.9 & 80.1 & 34.8 & 49.5 \\ 
 $\mathsf{META}$ & & \textbf{\color{red}{36.3}} & \textbf{\color{red}{35.1}} & \textbf{\color{red}{22.5}} & \textbf{\color{red}{49.9}} & \textbf{\color{red}{67.5}} & \textbf{\color{red}{86.1}} & \textbf{\color{red}{42.1}} & \textbf{\color{red}{57.0}} \\ \midrule
 SNR$^{\ast}$ \cite{jin2020style} & \multirow{6}{*}{\begin{tabular}[c]{@{}c@{}}Protocol-3\\ (Full Images)\end{tabular}} & 17.5 & 17.1 & 7.7 & 22.0 & 52.4 & 77.8 & 25.9 & 39.0 \\
 QAConv$_{50}$$^{\ast}$ \cite{qaconv} & &  32.9 & 33.3 & 17.6 & \textit{\color{blue}{46.6}} & 66.5 & \textit{\color{blue}{85.0}} & 39.0 & 55.0 \\
 M$^{3}$L (ResNet-50) \cite{metalearning}$^{\ast}$ & &  32.3 & 33.8 & 16.2 & 36.9 & 61.2 & 81.2 & 36.6 & 50.6 \\
 M$^{3}$L (IBN-Net50) \cite{metalearning}$^{\ast}$ & & 35.7 & 36.5 & 17.4 & 38.6 & 62.4 & 82.7 & 38.5 & 52.6 \\
 MetaBIN \cite{metanorm}$^{\ast}$ & & \textit{\color{blue}{43.0}} & \textit{\color{blue}{43.1}} & \textit{\color{blue}{18.8}} & 41.2 & \textit{\color{blue}{67.2}} & 84.5 & \textit{\color{blue}{43.0}} & \textit{\color{blue}{56.3}} \\ 
 $\mathsf{META}$ & & \textbf{\color{red}{47.1}} & \textbf{\color{red}{46.2}} & \textbf{\color{red}{24.4}} & \textbf{\color{red}{52.1}} & \textbf{\color{red}{76.5}} & \textbf{\color{red}{90.5}} & \textbf{\color{red}{49.3}} & \textbf{\color{red}{62.9}} \\ \bottomrule
\end{tabular}
}
\label{table:protocol2}
\end{table*}

\begin{table}[t]
\center
\setlength{\abovecaptionskip}{0.1cm}
\caption{Ablation study on the effectiveness of individual components and the design of global branch. The experiment is conducted under protocol-3. `C3', `MS', `M' are the abbreviations of the CUHK03, MSMT17, and Market1501 respectively. The best results are highlighted in bold.}
\resizebox{0.9\textwidth}{!}{
\begin{tabular}{lccccccaa}
\toprule
\multirow{2}{*}{Method} & \multicolumn{2}{c}{Target: C3} & \multicolumn{2}{c}{Target: MS} & \multicolumn{2}{c}{Target: M} & \multicolumn{2}{a}{Average} \\ 
& mAP & Rank-1 & mAP & Rank-1 & mAP & Rank-1 & mAP & Rank-1 \\ \midrule
 w/o global branch & 26.4 & 26.2 & 10.3 & 28.3 & 44.1 & 71.6 & 26.9 & 42.0 \\
 w/o expert branch & 33.6 & 33.7 & 20.5 & 45.8 & 71.9 & 87.6 & 42.0 & 55.7 \\
 w/o aggregation module & 46.0 & 45.5 & 23.3 & 50.9 & 75.1 & 89.6 & 48.1 & 62.0 \\\midrule
 BN-BN & 43.3 & 43.1 & 21.9 & 48.6 & 71.7 & 88.3 & 45.6 & 60.0 \\
 BN-IBN \cite{pan2018two} & 45.2 & 44.0 & 22.7 & 50.2 & 73.2 & 89.8 & 47.0 & 61.3 \\
 IN-IN & 41.5 & 40.2 & 18.7 & 46.0 & 68.3 & 86.7 & 42.8 & 57.6 \\
 \midrule
$\mathsf{META}$ & \textbf{47.1} & \textbf{46.2} & \textbf{24.4} & \textbf{52.1} & \textbf{76.5} & \textbf{90.5} & \textbf{49.3} & \textbf{62.9} \\ \bottomrule
\end{tabular}
}
\label{table:components}
\end{table}

\textbf{Implementation Details.} We resize all the images to $256 \times 128$. ResNet50 \cite{resnet} pretrained on ImageNet is used as our backbone. We set batch size to 64, including 16 identities and 4 images per identity. Similar to \cite{multiexpert}, we perform color jitter and discard random erasing for the data augmentation. We train the model for 120 epochs and adopt the warmup strategy in the first 500 iterations. The learning rate is initialized as $3e^{-4}$ and divided by 10 at the 40th and 70th epochs respectively. The margins $\alpha_1, \alpha_2$ in Eq.~(\ref{equ:consis}) are set to be 0.1.

\subsection{Comparison with State-of-the-art Methods}
\textbf{Comparison under protocol-1.} We compare our method with other state-of-the-arts under protocol-1, as shown in Table~\ref{table:protocol1}. We report some results of other methods which leverage DukeMTMC-reID \cite{duke} in the source domains, while we remove it from our training sets. Although we use fewer source domains, we still get the best performance. Specifically, from the results, we can find that $\mathsf{META}$ achieves the best performances on the PRID, GRID and VIPeR, while RaMoE \cite{multiexpert} gives the highest points on the iLIDs dataset. $\mathsf{META}$ significantly outperforms other methods by at least $2.0 \%$ and $2.3 \%$ in average mAP and Rank-1 respectively.

\textbf{Comparison under protocol-2 and protocol-3.} We compare our method with other state-of-the-arts under protocol-2 and protocol-3, as shown in Table~\ref{table:protocol2}. `Training Sets' denotes that only the training sets in the source domains are used for training and `Full Images' denotes that all images in the source domains ($i.e.$ including training and testing sets) are leveraged at training time. The results show that $\mathsf{META}$ outperforms other methods by a large margin on all the datasets and under both protocols. Specifically, $\mathsf{META}$ surpasses other methods, on average, by at least $4.6 \%$ mAP, $5.6 \%$ Rank-1 and $6.3 \%$ mAP, $6.6 \%$ Rank-1 under protocol-2 and protocol-3 respectively. The results have shown our model's superiority in domain generalization.

\subsection{Ablation Study}
\setlength{\parindent}{2em}\textbf{The effectiveness of the individual branches.} We study ablation studies on the effectiveness of individual branches, as shown in the first, second, and last rows of Table~\ref{table:components}. The experiment is conducted under protocol-3. We train our model without the global branch or expert branch for comparison. From the results, we can find that mAP drops $20.7 \%$, $14.1 \%$ and $32.4 \%$ on the CUHK03, MSMT17 and Market1501 respectively when the global branch is discarded. The mAP also drops $13.5 \%$, $3.9 \%$ and $4.6 \%$ on the CUHK03, MSMT17 and Market1501 respectively when the expert branch is discarded. The results have demonstrated the effectiveness of both the global and expert branches. Furthermore, we visualize the features extracted by different branches via t-SNE \cite{van2008visualizing}, as shown in Fig.~\ref{fig:visual}(a). Different colors denote various IDs. We find that the expert branch pushes features from different IDs away while the global branch pulls the features from same ID closer. Thus, both branches are integrated for better ReID performance.

\textbf{The effectiveness of aggregation module.} We study ablation studies on the effectiveness of aggregation module, as shown in the third and last rows of Table~\ref{table:components}. The experiment is conducted under protocol-3. \emph{`w/o aggregation module'} denotes that we remove the aggregation module and directly integrate multiple experts with FID. The results show that the aggregation module gives the performance gains of 1.1\%, 1.1\% and 1.4\% for mAP on CUHK03, MSMT17 and Market1501 respectively. The results have validated the effectiveness of the aggregation module for adaptively integrating diverse domain experts to mimic unseen target domain.

\setlength{\tabcolsep}{3.0pt}
\begin{table}[t]
\centering
\begin{minipage}{0.56\linewidth}
\centering
\footnotesize
\caption{Ablation study on the performance of the individual features under protocol-3.}
\resizebox{0.95\textwidth}{!}{$
\begin{tabular}{lcccccc}
\toprule
\multirow{2}{*}{Method} & \multicolumn{2}{c}{Target: C3} & \multicolumn{2}{c}{Target: MS} & \multicolumn{2}{c}{Target: M} \\
 & mAP & Rank-1 & mAP & Rank-1 & mAP & Rank-1 \\ \midrule
\emph{F-global} & 46.9 & 46.0 & 24.1 & 52.0 & 76.4 & 90.3 \\
\emph{F-exp} & 42.9 & 42.0 & 10.2 & 28.7 & 45.7 & 72.3 \\ \midrule
$\mathsf{META}$ & \textbf{47.1} & \textbf{46.2} & \textbf{24.4} & \textbf{52.1} & \textbf{76.5} & \textbf{90.5} \\ \bottomrule
\end{tabular}
$}
\label{table:feature}
\end{minipage}
\begin{minipage}{0.4\linewidth}
\centering
\footnotesize
\caption{Ablation study on loss functions under protocol-3.}
\resizebox{0.99\textwidth}{!}{$
\begin{tabular}{cccccc}
\toprule
\multirow{2}{*}{$\mathcal{L}_{base}$} & \multirow{2}{*}{$\mathcal{L}_{cross}$} & \multirow{2}{*}{$\mathcal{L}_{tri}$} & \multirow{2}{*}{$\mathcal{L}_{consis}$} & \multicolumn{2}{c}{Target: MSMT17} \\ 
& & & & mAP & Rank-1 \\ \midrule
$\surd$ & & & & 21.2 & 48.4 \\
$\surd$ & $\surd$ & & & 23.5 & 50.9 \\
$\surd$ & & $\surd$ & & 22.8 & 50.4 \\ 
$\surd$ & & & $\surd$ & \textbf{24.4} & \textbf{52.1} \\ \bottomrule
\end{tabular}
$}
\label{table:loss}
\end{minipage}
\end{table}
	
\textbf{The design of global branch.} The global branch is designed based on the findings from  \cite{pan2018two} that adding IN layers after BN layers could significantly improve the domain generalization performance of the model. We compare our design with other architectures of the global branch, as shown in the last four rows of Table~\ref{table:components}. The experiment is conducted under protocol-3. We respectively replace IN in the \emph{Global-IN} with BN and IBN \cite{pan2018two}, and replace BN in the \emph{Global-BN} with IN for comparison. The results show that our design achieve the best results, surpassing other architectures by $2.9\%$, $1.6\%$ and $5.3\%$ respectively in average Rank-1. The results have demonstrated the effectiveness of our design of global branch to help extract domain-invariant features.

\textbf{Performance of individual features.} We study ablation studies on the performance of individual features, as shown in Table~\ref{table:feature}. The experiment is conducted under protocol-3. We separately inference with \emph{F-global} and \emph{F-exp} for comparison. The results show that \emph{F-global} has a similar performance with $\mathsf{META}$ which concatenates \emph{F-global} and \emph{F-exp} for testing. We think the reason is that the expert branch is able to help the backbone extract more generalizable features, and therefore could improve the domain generalization performance of the global branch. As a result, it is feasible to only leverage the global branch during testing for faster inference.

\textbf{The effectiveness of loss function components.} We study ablation studies on the effectiveness of loss function components, as shown in Table~\ref{table:loss}. The experiment is conducted under protocol-3. $\mathcal{L}_{base}$ is defined in Eq.~(\ref{equ:base}) for training the global and expert branch. $\mathcal{L}_{cross}$ and $\mathcal{L}_{tri}$ indicate that we replace $\mathcal{L}_{consis}$ with cross-entropy loss and triplet loss respectively to update the aggregation module.  From the first and fourth rows, we can find that $\mathcal{L}_{consis}$ gives performance gains of $3.2 \%$ and $3.7 \%$ for mAP and Rank-1 accuracy respectively. From the last three rows, we can find that $\mathcal{L}_{consis}$ achieves the best performance, which surpasses $\mathcal{L}_{cross}$ and $\mathcal{L}_{tri}$ by $1.2 \%$ and $1.7 \%$ Rank-1 accuracy respectively. The results have demonstrated the effectiveness of our proposed $\mathcal{L}_{consis}$.

\begin{figure}[t]
\centering
\scriptsize
\setlength\tabcolsep{0mm}
\renewcommand\arraystretch{1.0}
\begin{tabular}{cc}
	\includegraphics[width=0.46\linewidth]{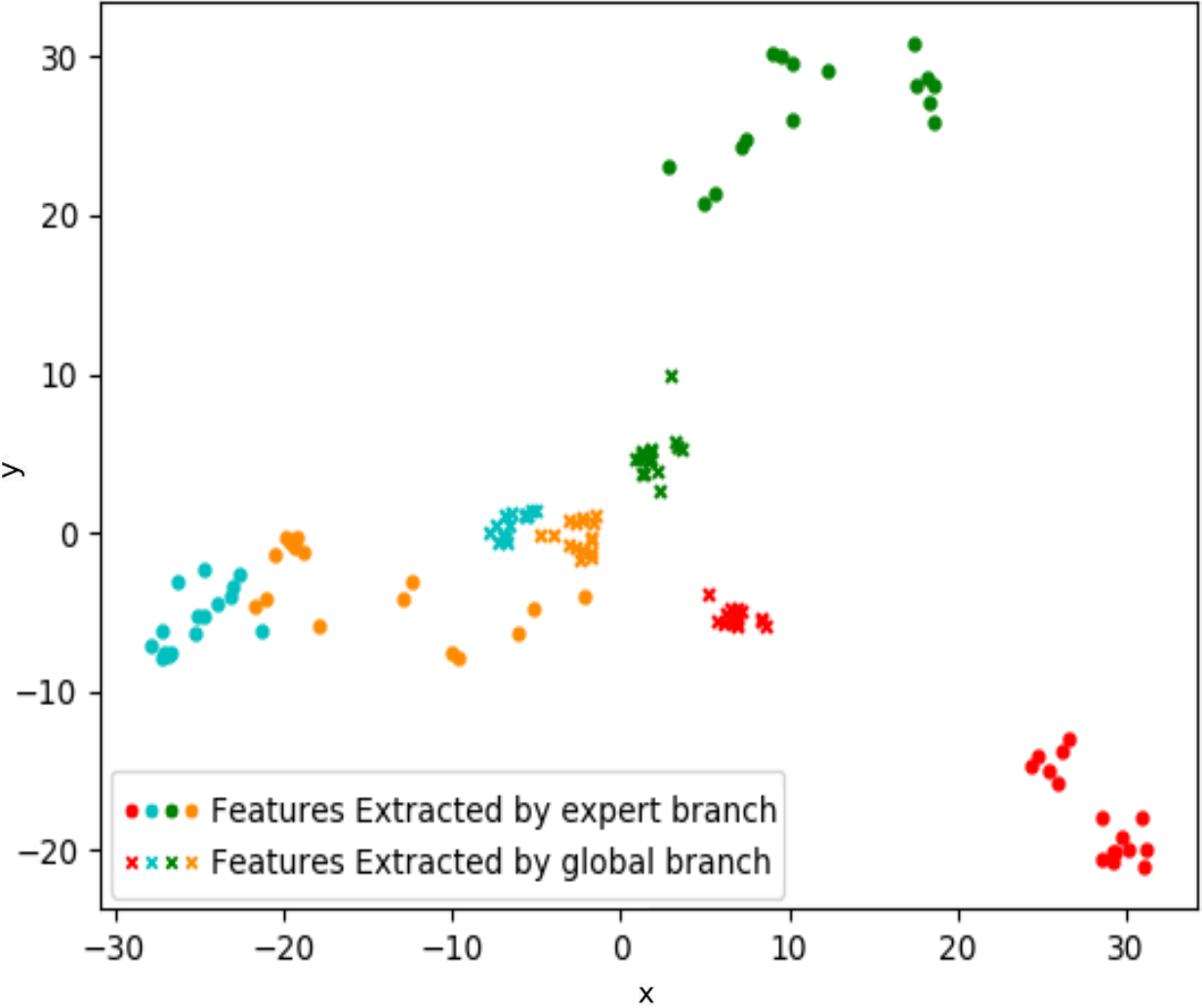} & 
	\includegraphics[width=0.50\linewidth]{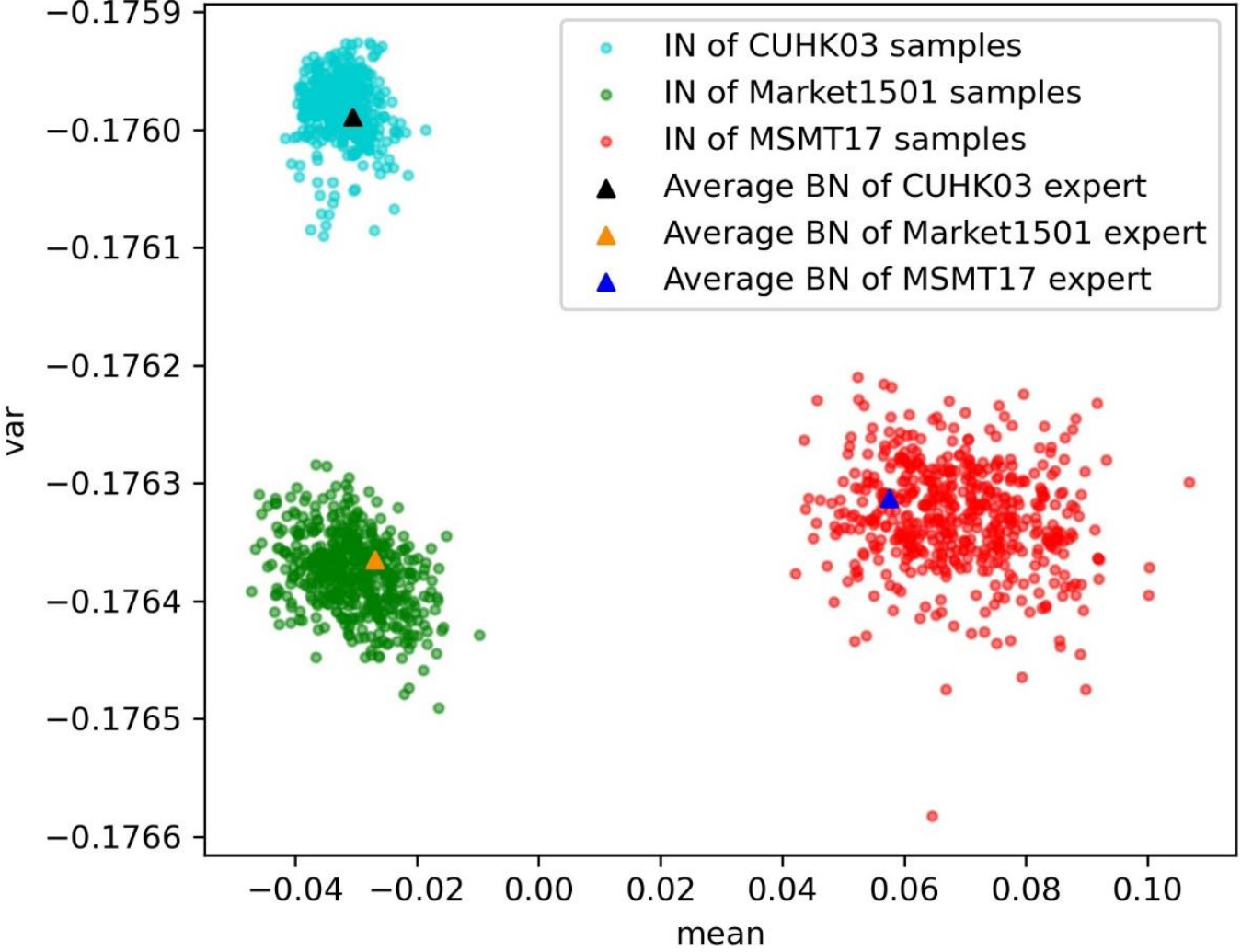}\\
	(a) Visualization of different branches & 
	(b) Visualization of statistics
\end{tabular}
\caption{(a) Visualization of the features extracted by different branches.  Various colors denote different IDs. (b) Visualization of statistics, the results illustrate the justification for calculating FID between BN and IN statistics.}
\label{fig:visual}
\end{figure}

\textbf{The justification for calculating
FID between BN and IN statistics.} Both BN and IN can be seen as approximations of different Gaussian distributions, thus we can simply adopt FID to measure the difference between them. We expect through our learning scheme, BN and IN statistics could reflect the properties of the source and target domain respectively. Fig.~\ref{fig:visual}(b) plots the average BN of multiple experts and IN statistics of samples from different domains via t-SNE \cite{van2008visualizing}. The horizontal and vertical axes represent the mean and standard deviation of the statistics respectively. The result shows that different domain clusters can be divided by their IN statistics. Additionally, IN statistics of the samples are closer to the average BN of the expert from the same domain. The result illustrates the justification for calculating FID between BN and IN statistics.

\section{Conclusion} This paper presents a new approach called Mimic Embedding via adapTive Aggregation ($\mathsf{META}$) for Domain generalizable (DG) person re-identification (ReID). $\mathsf{META}$ is a lightweight ensemble of multiple experts sharing all the parameters except for the domain-specific BN layers. Besides multiple experts, $\mathsf{META}$ leverages Instance Normalization (IN) and introduces it into a global branch to pursue invariant features across domains. Meanwhile, $\mathsf{META}$ develops an aggregation module to adaptively integrate multiple experts with the relevance of an unseen target sample w.r.t. source domains via normalization statistics. Extensive experiments demonstrate that $\mathsf{META}$ surpasses state-of-the-art DG ReID methods by a large margin. 

\section{Acknowledgement} The authors would like to thank reviewers for providing valuable suggestions to improve this paper. This work is supported by the National Natural Science Foundation of China (Grant No. U1836217) and the Beijing Nova Program under Grant Z211100002121108.
\clearpage
%
%
\bibliographystyle{splncs04}
\bibliography{eccv2022submissionCR}
\end{document}